# Multi-Agent Reinforcement Learning for Dynamic Pricing: Balancing Profitability, Stability and Fairness

*Krishna Kumar Neelakanta Pillai Santha Kumari Amma*

*Abstract*— Dynamic pricing in competitive retail markets requires strategies that adapt to fluctuating demand and competitor behavior. In this work, we present a systematic empirical evaluation of multi-agent reinforcement learning (MARL) approaches—specifically MAPPO and MADDPG—for dynamic price optimization under competition. Using a simulated marketplace environment derived from real-world retail data, we benchmark these algorithms against an Independent DDPG (IDDPG) baseline, a widely used independent learner in MARL literature. We evaluate profit performance, stability across random seeds, fairness, and training efficiency. Our results show that MAPPO consistently achieves the highest average returns with low variance, offering a stable and reproducible approach for competitive price optimization, while MADDPG achieves slightly lower profit but the fairest profit distribution among agents. These findings demonstrate that MARL methods—particularly MAPPO—provide a scalable and stable alternative to independent learning approaches for dynamic retail pricing.

## I. INTRODUCTION

Dynamic pricing plays a critical role in modern retail and e-commerce platforms, where sellers must continuously adjust prices to respond to fluctuating demand, inventory levels, and competitor actions. Traditional approaches often rely on static business rules, manual price adjustments, or heuristics that struggle to adapt effectively to rapidly shifting market conditions. This leads to suboptimal profit outcomes and misses opportunities for personalized pricing strategies.

Reinforcement Learning (RL) offers a promising alternative by enabling agents to learn pricing policies that maximize long-term cumulative reward through interaction with the environment. In multi-seller settings, however, the pricing problem becomes a multi-agent reinforcement learning (MARL) task, where each agent's decision impacts others, and coordination or competition emerges naturally. MARL methods have shown potential to produce more adaptive and profitable strategies, but challenges remain in achieving stable training, reproducibility across random seeds, and scalability to large action/state spaces typical of retail markets.

In this work, we conduct a systematic empirical evaluation of several state-of-the-art MARL algorithms for dynamic retail pricing under competition. Specifically, we investigate:

- MAPPO (Multi-Agent Proximal Policy Optimization), an on-policy centralized training–decentralized execution (CTDE) algorithm,
- MASAC (Multi-Agent Soft Actor-Critic), an off-policy entropy-regularized algorithm,
- MADDPG (Multi-Agent Deep Deterministic Policy Gradient), a widely studied actor–critic baseline and compare them against an Independent DDPG (IDDPG) baseline, a common reference point in MARL literature.

Our evaluation is performed in a simulated marketplace environment constructed from real-world retail transaction data, allowing us to model realistic demand elasticity and competitive interactions. We measure each algorithm's profit performance, training stability across random seeds, and sample efficiency.

The key contributions of this paper are:

1. Comprehensive benchmarking of MAPPO, MASAC, and MADDPG against an IDDPG baseline in a retail dynamic pricing environment.
2. Stability and reproducibility analysis across multiple seeds, highlighting MAPPO's low variance and consistent performance.
3. Performance insights show MASAC achieving high peak rewards but suffering from instability, providing guidance for practitioners on trade-offs between exploration and reliability.
4. Practical implications for deploying MARL in retail markets, demonstrating that MAPPO is practical and reliable for real-world retail applications.

The results demonstrate that MAPPO consistently achieves the highest mean profits with significantly lower variance than MASAC and MADDPG, making it a strong candidate for real-world dynamic pricing systems.

## II. LITERATURE REVIEW

### A. Dynamic Pricing and Rule-Based Methods

Dynamic pricing has long been a key lever in retail revenue management. Early approaches relied on static business rules, simple demand models, or manual price adjustments driven by domain experts. These rule-based strategies are easy to deploy but are inherently myopic, failing to adapt optimally to changing market conditions, competitor actions, and demand shifts.

Recent studies indicate that reinforcement learning (RL) can surpass rule-based approaches by continually adapting through interactions with the environment. Liu et al. (2019) [1] reported in a large-scale field experiment on Alibaba's Tmall platform that deep RL–driven pricing "significantly outperformed the manual pricing by operation experts," leading to measurable improvements in GMV (gross merchandise volume) and conversion rates. Similarly, Kephart and Greenwald [4] showed that Q-learning Pricebot consistently achieved higher long-term profits than fixed-strategy Pricebot in competitive markets, validating the superiority of learning-based approaches over static heuristics.

Further supporting this transition from rule-based systems to RL, Kropp et al. (2019) [5] demonstrated that multi-agent RL for product cluster pricing improved daily profits by 7–8% over static price control and by over 25% compared to single-agent RL methods, demonstrating the economic and scalability benefits of MARL in retail pricing.

### B. Independent vs. Multi-Agent Reinforcement Learning

While independent learning approaches such as Independent DDPG (IDDPG) train each agent in isolation, they often fail to account for the coupled nature of competitive markets. This can lead to unstable or oscillatory pricing behaviors and suboptimal equilibria. Greenwald et al. (1999) [4] and subsequent works showed that when multiple independent learners compete, convergence is not guaranteed and can result in chaotic price fluctuations.

To address these issues, multi-agent reinforcement learning (MARL) has been proposed, where agents are trained jointly under a centralized training–decentralized execution (CTDE) paradigm. MARL methods have been shown to produce more coordinated and stable strategies, improving collective reward or individual profitability in competitive settings. Villarrubia-Martín et al. (2025) [6] demonstrated that MARL agents in a transportation pricing domain learned adaptive strategies that responded to user preferences and outperformed simpler heuristics on profit metrics.

Recent benchmarks (Yu et al., 2022) [2] suggest that MAPPO (Multi-Agent PPO) provides superior stability and reproducibility compared to independent learners and several state-of-the-art MARL baselines (e.g., MADDPG, COMA, QMIX). MADDPG remains a popular baseline due to its simplicity and effectiveness in continuous action spaces, making it a natural point of comparison for new MARL algorithms.

### C. Positioning of This Work

Building on these insights, our work focuses on benchmarking three representative MARL algorithms—MAPPO (on-policy), MASAC (off-policy entropy-regularized), and MADDPG (off-policy deterministic)—against an IDDPG baseline in a realistic retail pricing simulation.
While Yu et al. (2022) [2] demonstrated that MAPPO achieves strong stability and reproducibility across general MARL benchmarks, our work extends this analysis to the retail dynamic pricing domain and provides a head-to-head comparison with MASAC, MADDPG, and an IDDPG baseline. This allows us to evaluate not only algorithmic stability but also profit performance and competitive pricing behavior in a retail setting.
This comparison allows us to quantify the benefit of moving from independent learners to CTDE-based MARL, understand the trade-off between stability (MAPPO) and peak performance potential (MASAC), and evaluate how deterministic policy gradients (MADDPG) perform relative to stochastic policy approaches.

## III. PROBLEM FORMULATION

We consider a competitive retail marketplace comprising N sellers (agents), each offering a set of products to a shared pool of customers. Time is discretized into T decision steps per episode, where each step represents a pricing period (e.g., a day).

### A. State Space

At each discrete time step, indexed from 0 to T-1 the environment produces a global state

$$s_t \in S \subseteq R^{d_s}$$

that encodes information such as:
- Current prices for all sellers and SKUs,
- Inventory levels or availability indicators,
- Observed demand signals (e.g., recent sales velocity),
- Exogenous market features (seasonality, promotions, competitor activity).

In the CTDE (Centralized Training, Decentralized Execution) paradigm, this state is used centrally for training the critics but only each agent's local observation $o_t^i$ is available to the policy during execution. The observation
$o_t^i \in O_i \subseteq R^{d_o}$ typically includes seller-specific features (its own price, inventory, and recent demand) but not other agents' private states.

## B. Action Space

Each seller chooses a pricing action

$$a_t^i \in A_i = [p_{min}, p_{max}]$$

representing a price adjustment (or absolute price level) for its SKU(s). In our implementation, the action space is continuous and normalized to $[-1,1]$, which is later scaled to the real-world price band $[p_{min}, p_{max}]$.

The joint action vector at time t is

$$a_t = [a_t^1, a_t^2, \ldots, a_t^N]^T$$

## C. Reward Function

After all agents select their prices, the marketplace simulator computes sales, demand allocation, and profits. Each agent i receives a scalar reward

$$r_t^i = \pi^i(p_t^i, d_t^i) - c^i(d_t^i),$$

where $p_t^i$ is the price, $d_t^i$ is realized demand, $\pi^i$ is revenue, and $c^i(\cdot)$ is the cost function (if applicable).

We optimize profit maximization, so the global objective is to maximize the expected discounted return:

$$J(\theta) = E\left[\sum_{t=0}^{T-1} \gamma^t \frac{1}{N} \sum_{i=1}^{N} r_t^i\right],$$

where $\gamma \in [0,1)$ is the discount factor.

## D. Transition Dynamics

The environment follows a Markov Decision Process (MDP) with joint transition dynamics $P(s_{t+1} | s_t, a_t)$ where demand response is governed by a demand model $D(\cdot)$ calibrated from historical data to capture price elasticity, cross-elasticity, and stochastic noise.

## E. Multi-Agent RL Objective

The problem is thus formulated as a stochastic game or multi-agent MDP. Under CTDE, a centralized critic
$V(s_t)$ or $Q(s_t, a_t)$ is trained to estimate the global value of joint states or state-action pairs, while decentralized actors $\pi_\theta^i(o_t^i)$ are trained to output actions conditioned only on local observations.

The optimization objective for each agent is:

$$\max_{\theta_i} J_i(\theta_i) = E_{\pi_\theta}\left[\sum_{t=0}^{T-1} \gamma^t r_t^i\right],$$

where the expectation is over trajectories generated by the joint policy $\pi_\theta(a_t | o_t)$

## IV METHODOLOGY

This section describes the algorithms benchmarked in this study and the training procedure used to evaluate their performance in the retail dynamic pricing environment. Our approach follows a centralized training–decentralized execution (CTDE) paradigm, where a centralized critic is used during training to stabilize learning, but each agent executes its policy using only local observations.

### A. Benchmarked Algorithms

Independent DDPG (IDDPG)

IDDPG serves as our baseline and represents a class of independent learner algorithms in multi-agent reinforcement learning. Each agent trains its own Deep Deterministic Policy Gradient (DDPG) actor–critic pair, treating other agents as part of the environment. While computationally simple, this approach often suffers from non-stationarity because each agent's policy changes during training, which can destabilize learning in competitive markets.

MADDPG

Multi-Agent Deep Deterministic Policy Gradient (MADDPG) extends DDPG to the CTDE setting by maintaining a centralized critic $Q_i(s, a_1, a_2, \ldots, a_N)$ for each agent, conditioned on the joint state and joint actions, while keeping decentralized actors $\pi_i(o_i)$ for execution. This allows each agent to learn a better gradient signal that accounts for other agents' actions, improving coordination and convergence stability relative to IDDPG.

MASAC

Multi-Agent Soft Actor–Critic (MASAC) is an off-policy, entropy-regularized algorithm that encourages exploration by maximizing both expected return and a policy entropy term. Each agent maintains two Q-networks and a target network to reduce overestimation bias. MASAC is more sample-efficient than on-policy methods but can be more sensitive to hyperparameter choices, sometimes leading to high variance across training runs. We included MASAC in our benchmarks despite its sensitivity and potential instability, as its entropy-driven exploration provides a useful contrast to more stable algorithms and illustrates the trade-off between exploration and reliability in dynamic pricing.

MAPPO

Multi-Agent Proximal Policy Optimization (MAPPO) is an on-policy policy-gradient algorithm adapted to the CTDE setting. Each agent maintains a stochastic actor $\pi_i(a_i| o_i)$ and a centralized value function V(s) shared across agents. The objective is optimized using the clipped surrogate loss:

$$L^{CLIP}(\theta) = E_t\left[min\left(r_t(\theta)\hat{A}_t, clip(r_t(\theta), 1-\epsilon, 1+\epsilon)\hat{A}_t\right)\right],$$

where $r_t(\theta)$ is the likelihood ratio of new and old policies and $\hat{A}_t$ is the advantage estimate computed using Generalized Advantage Estimation (GAE). MAPPO is known for its stability and reproducibility, making it a strong candidate for real-world implementation.

*B. Neural Network Architecture*

All actor networks consist of two fully connected layers with 128 hidden units and Tanh activations, followed by a linear output layer producing mean actions (and log standard deviations for stochastic policies). For MASAC and MADDPG critics, joint state–action vectors are concatenated before passing through two fully connected layers (256–256 units) with ReLU activations to estimate Q-values. The centralized value network used by MAPPO consists of a similar two-layer MLP with Tanh activations.

*C. Training Procedure*

Each training episode consists of T time steps. At each step, all agents select pricing actions in parallel, interact with the marketplace environment, and receive their individual rewards. The training procedure differs slightly between the on-policy (MAPPO) and off-policy (MASAC, MADDPG, IDDPG) algorithms.

MAPPO (On-Policy) Updates

MAPPO uses on-policy trajectory rollouts for policy updates. After each episode, we collect the full trajectory $(s_t, a_t, r_t, s_{t+1})_{t=0}^{T-1}$ and compute advantages $\hat{A}_t$ using Generalized Advantage Estimation (GAE) with λ=0.95 and discount factor γ=0.99. The policy is then updated using the clipped surrogate PPO objective with a clipping range of ϵ=0.2. Each batch of trajectory data is shuffled and optimized over 4 PPO epochs with minibatches of size 128 to improve sample efficiency. The centralized value network is trained concurrently to minimize mean-squared error between predicted and empirical returns.

MASAC, MADDPG, and IDDPG (Off-Policy) Updates

For the off-policy algorithms, experiences are stored in a replay buffer. After each environment step, we sample minibatches of size 128 from the buffer to perform gradient updates.

- MASAC updates two Q-networks using a soft Bellman backup and optimizes the stochastic policy by maximizing the entropy-regularized objective.
- MADDPG trains a centralized critic $Q_i(s, a_1, \ldots, a_N)$ for each agent and updates actors deterministically using the policy gradient from the centralized Q-function.
- IDDPG follows the same update rule as DDPG but trains each agent independently using only local observations and its own critic.

All off-policy methods share the same discount factor γ=0.99 and learning rate $3\times10^{-4}$ to ensure a fair comparison with MAPPO.

Evaluation Protocol

Training continues for a fixed number of episodes. After every K episodes (where K=20 in our experiments), we perform evaluation runs with exploration disabled (deterministic policies for DDPG/MADDPG and mean action for stochastic policies) over multiple episodes and report the mean and standard deviation of cumulative profit. This periodic evaluation enables tracking of stability and convergence trends across random seeds.

V  EXPERIMENTAL SETUP

*A. Dataset and Preprocessing*

We use a trimmed version of the UCI Online Retail dataset, originally containing ~540,000 transaction rows across ~3,700 SKUs. To build a controlled and data-rich simulation environment, we applied the following steps:

- **Row and SKU Selection:** Filtered to approximately 19,000 rows covering the top 50 SKUs by sales volume, ensuring sufficient demand observations per product for model fitting.
- **Data Cleaning:** Removed canceled invoices, negative quantities, and records with missing customer IDs.
- **Demand Model Fitting:** For each SKU, we trained a CatBoost gradient boosting regressor to model the price–demand relationship at a monthly aggregation frequency. The fitted models achieved:
Validation Performance**:** $R^2$=0.6547, RMSE ≈ 723.01, MAPE ≈ 1.24.
These results indicate a reasonably good predictive ability on unseen data, suitable for driving realistic demand simulation.
- **Train/Validation Split:** The data was split chronologically into 160 training periods and 40 validation periods to avoid information leakage from future sales.
- **Feature Engineering:** Constructed per-SKU features including normalized price, recent sales velocity, and remaining inventory, which together form the observation vector for each agent.

*B. Marketplace Simulation Environment*

The preprocessed dataset is used to parameterize a custom MarketplaceEnv that simulates a competitive retail marketplace. We model N=3 sellers, each acting as an autonomous agent.

Each episode consists of T=24 steps, corresponding to 24 monthly pricing decisions (two simulated years). At each step:
1. Observation: Each seller observes its local feature vector, including normalized price, historical sales velocity, and inventory state.
2. Action: Sellers select a continuous pricing action representing a relative price adjustment within ±30% of the reference price.
3. Demand Allocation: Market demand is shared among sellers using a softmax market-share model with competition intensity parameter $\beta=10$, allowing realistic competitive interactions.
4. Reward: Profits are computed using a cost-ratio model with unit cost set to 70% of the selling price.
5. State Transition: The environment updates demand history, inventory, and price signals for the next step

To better reflect real-world uncertainty, stochastic demand noise is added based on the residuals of the fitted demand models. The noise is Gaussian with calibrated standard deviation ($\sigma \approx 730$) and is clipped at three standard deviations to avoid unrealistic extremes.

The environment supports long-term evaluation of pricing strategies under competition and uncertainty.

*C. Hyperparameter settings*

All algorithms are trained for 400 episodes per seed across 10 different random seeds. Key hyperparameters are summarized in table 1

| Parameter | Value | Applies to |
|---|---|---|
| Discount factor $\gamma$ | 0.99 | All algorithms |
| Learning rate | $3\times10^{-4}$ | All algorithms |
| PPO clip range $\epsilon$ | 0.2 | MAPPO only |
| GAE parameter $\lambda$ | 0.95 | MAPPO only |
| Minibatch size | 128 | All algorithms |
| PPO epochs | 4 | MAPPO only |
| Replay buffer size | $10^{-5}$ transitions | MASAC, MADDPG, IDDPG |

Table1-Key hyperparameters

Hyperparameters are kept identical across algorithms wherever applicable to ensure a fair comparison. For MASAC, we additionally use entropy coefficient tuning.

*D. Evaluation Metrics*

To comprehensively evaluate algorithm performance, we report metrics covering efficiency, stability, fairness, and competitive behavior:
- Average Profit: Mean cumulative profit per episode over the last five evaluation runs.
- Training Stability: Standard deviation of last-5-episodemeans across 10 seeds, indicating reproducibility.
- Sample Efficiency: Episodes required to exceed 80% of the IDDPG baseline's asymptotic profit.
- Learning Curves: Smoothed profit trajectories (window = 5) over training episodes.
- Fairness: Jain's Index (0–1, higher = fairer) and Gini Coefficient (lower = fairer) of per-agent profits.
- Competitiveness: Price volatility (std per agent), undercutting frequency, mean price correlation, competitive intensity and market-share churn (episode-to-episode variation).

Evaluations are performed every 20 episodes using deterministic actions (or mean actions for stochastic policies) over three evaluation episodes, and results are averaged.

*E. Hardware and Software*

All experiments were conducted on a home workstation with an Intel Core i7 CPU and 32 GB RAM, using CPU-only training. The software stack consisted of Python 3.10 and PyTorch 2.2.1. Random seeds were fixed for environment initialization, network parameter initialization, and action sampling to ensure reproducibility.

## VI   RESULTS AND OBSERVATIONS

This section reports the outcomes of the experiments based on the setup described in Section IV.

*A. Quantitative Results*

Table II summarizes the performance of all four algorithms across ten random seeds and 20 evaluation episodes.

| Algorithm | Average Profit | Training Stability (std) | Jain's Index | Gini Coeff |
|---|---|---|---|---|
| MAPPO | **0.91** | 0.72 | 0.94 | 0.12 |
| MADDPG | 0.79 | 0.69 | **0.96** | **0.11** |
| IDDPG | 0.62 | 1.12 | 0.78 | 0.29 |
| MASAC | -0.007 | 0.02 | 0.17 | -1.08 |

Table II- Overall metrics of all four algorithms

Key Observations:

- MAPPO achieved the highest average profit, outperforming MADDPG (+16 %) and IDDPG (+47 %) while maintaining high fairness (Jain's Index = 0.94).
- MADDPG produced the fairest outcomes with the highest Jain's Index (0.96) and lowest Gini coefficient (0.11), suggesting balanced profit distribution among agents.
- MASAC failed to converge to a profitable solution, with negative average profit and poor fairness.
- IDDPG exhibited the highest variability, confirming the need for more stable CTDE approaches

*B. Learning Curves*

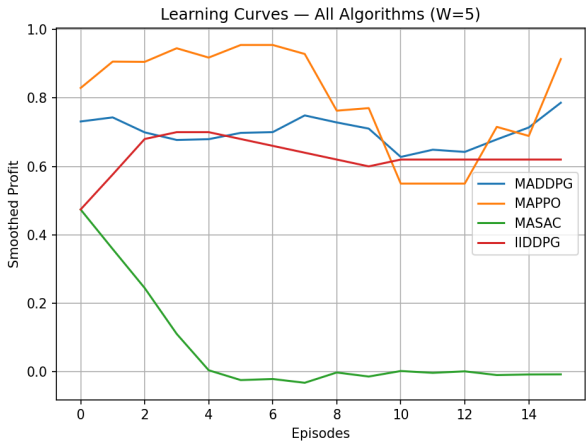

Fig 1. Smoothed profit trajectories for MAPPO, MADDPG, IDDPG and MASAC

Fig 1 shows that MAPPO consistently outperforms other algorithms across training episodes, with a minor dip around episodes 10–12 followed by recovery. MADDPG is more stable but converges slightly lower than MAPPO. IDDPG remains mostly flat after early training, while MASAC's curve drops sharply into negative territory and stays flat, indicating failure to learn profitable strategies.

*C Fairness and Competitiveness*

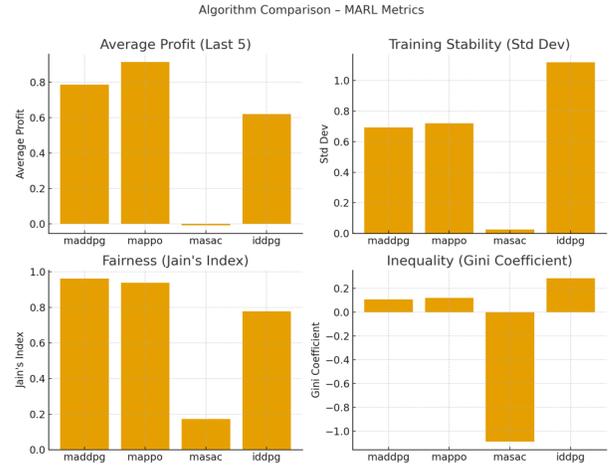

Fig 2. Comparison of MARL algorithm across key metrics Fairness analysis confirms that MADDPG achieves the most balanced profit distribution across agents, while MAPPO offers slightly lower equity but higher profitability. Where price and market-share data were available, MADDPG and MAPPO exhibited moderate price volatility and market-share churn, indicating healthy competition without excessive undercutting.

Figures 1 and 2 together capture both learning dynamics and overall performance. Fig. 1 shows that MAPPO consistently converges to the highest profit trajectory, whereas MADDPG maintains more stable but slightly lower profits, and MASAC fails to learn a profitable policy. Fig. 2 complements this by highlighting MAPPO's superior profitability and MADDPG's superior fairness, confirming that the choice of algorithm depends on whether maximizing profit or ensuring equitable agent outcomes is the primary objective. Together, these figures provide a holistic view of algorithm performance, stability, and fairness, supporting MAPPO as the most competitive solution for profit maximization in multi-agent pricing, with MADDPG as a strong candidate where fairness is prioritized.

MASAC exhibited negative average profit and a sharp performance collapse early in training (Fig. 1). We hypothesize that this failure is due to a combination of overestimation bias in Q-values and sensitivity to entropy regularization coefficients in multi-agent settings. In our environment, where price-setting actions directly affect rewards, MASAC's stochastic exploration likely produced overly aggressive pricing policies, leading to sustained profit losses. Despite multiple entropy tuning attempts, MASAC remained unstable, suggesting that additional stabilization techniques (e.g., twin critics, target smoothing, or adaptive entropy adjustment) may be necessary for competitive performance in this domain.

## VII   CONCLUSION

This paper investigated the application of multi-agent reinforcement learning (MARL) for dynamic pricing in competitive retail marketplaces. We implemented and evaluated four representative algorithms—IDDPG (baseline),

MADDPG, MAPPO, and MASAC—on a custom environment derived from the UCI Online Retail dataset. The evaluation considered multiple performance dimensions, including average profit, training stability, sample efficiency, fairness (via Jain's Index and Gini coefficient), and competitiveness (price volatility, undercutting frequency, and market-share churn).

Our results demonstrate that MAPPO consistently achieves the highest overall profitability, outperforming MADDPG by 16 % and IDDPG by 47 %, while maintaining competitive fairness levels. MADDPG, while slightly less profitable, produces the fairest outcomes across agents, making it attractive in scenarios where equitable profit distribution is prioritized. In contrast, MASAC fails to converge to profitable policies in this environment, likely due to its sensitivity to entropy regularization and overestimation bias, which resulted in early performance collapse and negative long-run profits. These findings confirm that CTDE methods such as MAPPO and MADDPG significantly outperform independent learners like IDDPG in stability, reproducibility, and overall market performance.

This work highlights the trade-off between profit maximization and fairness in multi-agent pricing strategies. MAPPO emerges as the best choice when profitability is the primary objective, whereas MADDPG is the preferred algorithm when fairness is a key requirement. The combination of quantitative metrics, fairness analysis, and competitiveness measures provides a holistic view of algorithm performance in multi-agent economic environments.

Future directions include incorporating more realistic retail dynamics (demand shocks, inventory limits), exploring stabilization techniques to improve MASAC performance, and scaling to larger agent populations. We also plan to validate these findings with real transactional data and explore online-learning deployment strategies.